\documentclass{article}

\usepackage{PRIMEarxiv}

\usepackage[utf8]{inputenc} 
\usepackage[T1]{fontenc}    
\usepackage{hyperref}       
\usepackage{url}            
\usepackage{booktabs}       
\usepackage{amsfonts}       
\usepackage{nicefrac}       
\usepackage{microtype}      
\usepackage{lipsum}
\usepackage{fancyhdr}       
\usepackage{graphicx}       
\graphicspath{{media/}}     

\title{Identification of Bias Against People With Disabilities in Sentiment Analysis and Toxicity Detection Models}

\author{
  Pranav Narayanan Venkit, Shomir Wilson \\
  College of Information Science and Technology\\
  Pennsylvania State University\\
  \texttt{\{pranav.venkit, shomir\}@psu.edu} \\
}
\begin{document}
\maketitle

\begin{abstract}
Sociodemographic biases are a common problem for natural language processing, affecting the fairness and integrity of its applications. Within sentiment analysis, these biases may undermine sentiment predictions for texts that mention personal attributes that unbiased human readers would consider neutral. Such discrimination can have great consequences in the applications of sentiment analysis both in the public and private sectors. For example, incorrect inferences in applications like online abuse and opinion analysis in social media platform can lead to unwanted ramifications, such as wrongful censoring, towards certain populations. In this paper, we address the discrimination against people with disabilities, PWD, done by sentiment analysis and toxicity classification models. We provide an examination of sentiment and toxicity analysis models to understand in detail how they discriminate PWD. We present Bias Identification Test in Sentiments (BITS), a corpus of 1,126 sentences designed to probe sentiment analysis models for biases in disability. We use this corpus to demonstrate statistically significant biases in four widely used sentiment analysis tools (TextBlob, VADER, Google Cloud Natural Language API and DistilBERT) and two toxicity analysis models trained to predict toxic comments on Jigsaw challenges (Toxic comment classification and Unintended Bias in Toxic comments). The results show that all exhibit strong negative biases on sentences that mention disability. We publicly release BITS Corpus for others to identify potential biases against disability in any sentiment analysis tools and also to update the corpus to be used as a test for other sociodemographic variables as well. \end{abstract}

\keywords{Ethics in AI \and Natural Language Processing \and Model Evaluation}

\section{Introduction}

Bias in natural language processing (NLP), and its consequences, have recently received substantial attention \cite{bolukbasi2016man, kiritchenko2018examining, caliskan2017semantics}. NLP models are frequently used to understand sentiment in social media platforms \cite{bisong2019google}. However, these models most often exhibit biases that lead to unintended discrimination towards specific population groups. An example for this was shown by \cite{caliskan2017semantics} where applying machine learning techniques to ordinary human language results in human-like semantic biases. This results in false positive predictions of texts as toxic or negative, based on certain group identifiers \cite{kennedy2020contextualizing}. 

According to the World Report on Disability by WHO, approximately one billion people, or 15\% of the word's population, experience some form of disability and almost everyone will be temporarily or permanently impaired at some point in their life \cite{bickenbach2011world}. Even though this understanding is well established today, people with disabilities (PWD) are still subjected to marginalization \cite{whittaker2019disability}. From a social lens, the term `Social Model of Disability' was coined by \cite{oliver1984politics} to bring aware and alleviate the social and economic barriers discrimination of PWD caused by cultural attitudes present in a social environment \cite{oliver1984politics}. Unfortunately the prejudice is still prevalent today \cite{chen2020disability} and the use of AI as solution has not helped \cite{whittaker2019disability}.

Discrimination in disability has only begun to get popular within the NLP community and is still under-explored \cite{hutchinson2020social}. The understanding of such implicit biases is essential in the field of NLP because numerous models are increasingly being developed as social solutions, such as fighting online abuse, measuring sentiment and toxicity, and creating classifiers that understand how group opinions work in a conversational platform \cite{noever2018machine, do2019jigsaw, pavlopoulos2020toxicity}. Amongst them, sentiment analysis has gained popularity in computational approach to understand attitude, opinions and hate speech in texts \cite{diaz2018addressing, pang2009opinion, del2017hate}. If such models contain implicit biases against disability, it would curb the freedom of speech and expression thus catalysing already existing inequalities on minorities. The result of such AI prejudice can provide misrepresentation in disability and age related political and social topics such Medicare and Social Security \cite{diaz2018addressing}.


We show bias in sentiment and toxicity analysis models based on scores generated by discussions in social media platforms i.e., Reddit and Twitter, around PWD. The results show that both social media based sentences are significantly rated more negative or toxic when associated to words related to disability. We also present a model-independent procedure for identifying sociodemographic bias in the same sentiment analysis tools, specifically along the dimensions of disability. We construct a corpus, termed the Bias Identification Test in Sentiments (BITS Corpus), that facilitates a bias test to analyse sentiment models for implicit discrimination against specific sociodemographic groups. The corpus consists of sentiment-neutral and sentiment-containing English sentences that probe a model to demonstrate latent biases against sociodemographic groups of interest. The corpus can also be modified and updated to act as a unique bias test for other sociodemographic factors beyond those we examined in this paper.

We examine four popular sentiment analysis tools: VADER \cite{gilbert2014vader}, TextBlob \cite{loria2018textblob}, Google Cloud Natural Language API and DistilBERT \cite{sanh2019distilbert}, and two popular toxicity analysis tools of the Detoxify library \cite{Detoxify}. Through the textual analysis task of the public tools, using BITS Corpus, we demonstrate the utility of the corpus and also identify biases in them. The tools selected for analysis are part of the well-known python libraries for NLP tasks, such as Pytorch, TextBlob and NLTK \cite{lin2018sentiment,rao_2019,monkeylearnblog_2019}. They are widely used, leading to potential biases in systems where they are deployed. The results show that all the four sentiment analysis tools show statistically significant bias towards PWD. Similar negative implicit bias appears in the toxicity classification tools as well.

\section{Related Work}

AI is rapidly being used in various socio-economic fields to provide decisions and classifications of huge resources. An AI system models the world based on the data it has been fed \cite{whittaker2019disability}. Hence, not including a specific population or dataset completely excludes those entities from identification or classification. Such biased system cause huge ramifications in any social environment \cite{ marshall2019uber, whittaker2019disability}. The work done by \cite{guo2020toward} and \cite{whittaker2019disability} showcases how various AI application, like Computer Vision and Text Processing, demonstrate bias against PWD as it is trained on data largely containing the non-disabled population. 

Bias is also largely found in NLP models as they handle large amounts of textual data. It has been shown that systems trained on textual data demonstrate human like biases easily \cite{caliskan2017semantics, bolukbasi2016man}. Texts and social media comments tend to contain a significant amount of vulgarity and hate \cite{do2019jigsaw}. Such skewed dataset causes AI models to be discriminatory towards certain subjects. This is seen in the various works done for sociodemographic bias identification in textual analysis \cite{sap2019risk, hutchinson2020social, guo2020toward, kiritchenko2018examining, kennedy2020contextualizing, park2018reducing}. In this work, we were motivated to check if textual data related to the language around PWD can lead to bias against them if used in a sentiment and toxicity analysis classifier that is built or defined around the non-disabled population.


People with disabilities, PWD, have been subjected to historical and present-day marginalization as well \cite{whittaker2019disability}, like other underrepresented population.
Literature also suggests that hate crime against disability is often hidden and is the most under-reported as compared to other forms of sociodemographic hate crimes \cite{macdonald2017disability, creese2014hate}. As mentioned by \cite{trewin2018ai}, `Disability is not a simple variable with a small number of discrete values. It has many dimensions and people can experience multiple disabilities' \cite{trewin2018ai}. It is easy for an AI application to easily discriminate PWD due to over-simplification of their model leading them to not understand such nuances \cite{o2016weapons}. Due to this, PWD do not fit under the category of the `normal' or the `majority'. A lot of work focus on understanding the discrimination caused by NLP models with related to race and gender \cite{bolukbasi2016man, kiritchenko2018examining, caliskan2017semantics, sheng2019woman, kurita2019measuring}, but research on disability is relatively unexplored \cite{hutchinson2020social, whittaker2019disability}.

Sentiment analysis is an important element in the architecture of many NLP systems \cite{medhat2014sentiment}. Many have tried to understand the performance difference of various public sentiment models \cite{ribeiro2016sentibench, kiani2018comparison} but not many have explored how they are biased. These models are widely used for various textual analysis and if such models are skewed, so will the result of the operation it is used in \cite{diaz2018addressing}. \cite{thelwall2018gender} tries to depict the gender bias associated to a few sentiment classifiers to know how they score reviews written by both the male and female gender. \cite{diaz2018addressing} works with understanding how these sentiment model can be biased towards age-related terms. Their works explores age related bias in 15 sentiment models and 10 widely used GLoVe embeddings. \cite{kiritchenko2018examining} provides a more defined analysis on how to examine the bias associated to various sentiment systems. They analysed 218 sentiment classifiers that were submitted to SemEval 2018 Task 1 \cite{mohammad2018semeval}. The work presented by \cite{kiritchenko2018examining}, or similar works, did not focus the population related to PWD. 

In the field of NLP, toxicity is a parameter that is recently being explored in social media platforms \cite{do2019jigsaw}. This parameter has been used to check if social comments are toxic, to improve online behaviour without human intervention \cite{jigsaw_2021, jigsaw2021, jigsaw21}. The issue with such analysis is that it tends to flag non-toxic words related to people with disability as toxic as well \cite{hutchinson2020social}. Comments written by the PWD community can be censored exacerbating the already reduced visibility of disability in the public discourse \cite{hutchinson2020social}. 

The language used by the PWD community can lead to potential negative discrimination against them. People with autism may express emotions differently in writing than the neurotypical population. Such languages can lead to misclassification of their emotional state or personality \cite{guo2020toward}. Textual analysis done by \cite{hutchinson2020social} on comments related to PWD show how high negative sentiment words such as gun violence, homelessness and drug additions are over-represented in texts discussing mental illness. These associations may not be surprising but using social AI solutions with these skewed results can significantly shape online conversations as well as provide wrongful analysis of the PWD community. Bias against disability is not well studied in well known sentiment analysis and text analysis libraries that are used to make decisions in politics, finance, employment and education \cite{kuncel2014hiring, miller2015can, rubel2016black, diaz2018addressing}. As mentioned by \cite{guo2020toward}, `if these metrics are used as input to an automatic hiring system or automatic essay grading systems used with many standardized aptitude tests, text analysis systems can have accuracy and fairness challenges for people with cognitive and/or intellectual disabilities' \cite{guo2020toward}. PWD community are more likely to talk about disability, this could impact their opportunity to participate equally in online social fora, reducing their autonomy and dignity. Disability is often less about physical or mental impairment than it is about how society responds to impairments. Therefore we must be mindful of such NLP solutions used to represent them.

To look further into the issue, we analyse bias present in mentioned sentiment classification models using language and conversations pertained to PWD, in online social platforms. The bias of the selected tools will be shown by the difference in performance of terms surrounding PWD and neurotypical population. The use of social media data will show the language that is used by and around the PWD community in a social media space, providing context to the analysis. Following this we describe the creation of a corpus, which encompasses independent analysis of bias against PWD. We publicly release BITS Corpus, consisting of 1,126 sentences. The template of the corpus is motivated by the structure followed by the Equity Evaluation Corpus (EEC) \cite{kiritchenko2018examining}. The difference in our solution lies in how BITS Corpus is intended to be used to identify the possible presence of multiple sociodemographic biases, i.e., bias towards PWD, gender and race, independently. The templates present in the test can be modified for checking bias related to any other sociodemographic group as well. The goal is to create a solution that can be used by all to check if a sentiment model discriminates data based on a wider group. We also demonstrate the presence of these biases in public sentiment and toxicity analysis models as they are widely used and easily accessible by many in the field of textual analysis.

\section{Social Media Analysis}

\begin{table}
\small
\centering
\begin{tabular}{l r}
\hline
\textbf{Sentence} & \textbf{Score} \\[0.5ex] 
\hline
My neighbour is a tall person. & 0.00\\
My neighbour is a beautiful person. & 0.85\\
My neighbour is a black person. & -0.16\\
My neighbour is a mentally handicapped person. & -0.10\\
My neighbour is a blind person. & -0.50\\
\hline
\end{tabular}
\caption{Example of TextBlob sentiment analysis library with a sentence set. It is seen that certain terms generate negative sentiments irrespective of context. }\label{example}
\end{table}

The need for a bias identification method for sentiment analysis model needs to be highly consider, especially with respect to PWD. As a simple illustration, used TextBlob on a series of simple sentences. The scores generated for each sentence are shown in Table \ref{example}. We can clearly see that the scores are discriminatory to certain words. To gain a better understanding of this bias, we will look into how such sentiment analysis tools perform with social media comments.

There are many known public sentiment analysis tools and methods that is use to evaluate the sentiments of a text. This functionality has become very popular in several analytic platforms, especially with web and social media data \cite{ribeiro2016sentibench}. Due to this increase in usage, there have now been many libraries that are offer to measure the sentiment, toxicity and polarity of a sentence. To ease this task, several of developers have now made public libraries and tools that can be used by anyone, to create a sentiment analysis and identification model \cite{gilbert2014vader}. But these models are treated as `black boxes' and are used without understanding the presence of potential bias in them \cite{o2016weapons}.

In the initial analysis, we will showcase the possible biases present in three models that are used for sentiment analysis: VADER (Valence Aware Dictionary for sEntiment Reasoning), TextBlob Python library and DistilBERT. VADER is a lexicon and rule-based sentiment analysis tool that is specifically attuned to sentiments expressed in social media \cite{gilbert2014vader}. This is a rule-based sentiment analysis tool that has been shown to be especially effective for social media posts \cite{gilbert2014vader}. Textblob is an NLTK based python library that provides a simple function for fundamental NLP tasks such as part-of-speech tagging, sentiment analysis and classification. Finally, DistilBERT \cite{sanh2019distilbert} is a small, fast and light Transformer model trained by distilling BERT base algorithm \cite{devlin2018bert}. We will also look into two Toxicity Analysis libraries, released by Unitary. The Toxicity Classification libraries \footnote{https://huggingface.co/unitary/toxic-bert} are a high performing neural network based model that is trained on the Kaggle dataset that was published in the Toxic Comment and Jigsaw Unintended Bias in Toxicity Classification competition. We will term them as `Toxicity\_Original' and `Toxicity\_Biased' respectively. The scores for all the selected sentiment analysis models are standardized to provide a result from -1 (maximum negative sentiment) to +1 (maximum positive sentiment). The toxicity classification models score from -1 (maximum toxicity) to 0 (No toxicity).

To completely understand the presence of bias in these tools, we will be examining sentences that are used while discussing PWD. The results obtained from such posts will provide an understanding of bias that is present in social forums and models. For this analysis we will be looking into Twitter and Reddit platforms. We will follow an approach similar to \cite{diaz2018addressing} who used Reddit discussions to examine the presence of age related bias in sentiment analysis models. 

\subsection{Social Media Data Collection}

From the Reddit discussion forum, we looked into 238 blog posts and 1782 comments from the `Disability' community that talks about news, resources and perspectives pertaining to PWD. The conversations span from July 12, 2020 to July 15, 2021. For Twitter related conversations, we retrieved recent tweets containing any of the following terms or hashtags: `disability' , `disabled' , `\#disability'. We obtained a total of 53,454 tweets from 9th July, 2021 to  16th July, 2021. To examine bias in the models, we filtered unique simple sentences that talk about topics related to disability. For simplicity of analysis we did not consider sentences that mentioned topics related to very specific disabilities. We omitted sentences that were large conversations consisting of too many themes as well as sentences that had the term `abled' or themes related this word. Out of the complete collection, 70 and 141 sentences were selected from Reddit and TWitter dataset respectively.  

To understand how each model treats PWD as compared to people without disability, we replace the word `disability' and `disabled' from the selected social media sentences with words describing other groups. Through this procedure, we analyse language pertained to four groups, i.e. \textit{People with Disability}, \textit{People With Disability: Social Language}, \textit{People Without Disability} and \textit{Normalized Adjectives}. The group breakdown and their corresponding terms are shown in Table \ref{disability-group}. For shorter representation in tables, we will be using \textit{DSBL, DSBL:S, NDSBL} and \textit{NRMA} label tags.

The first group, \textit{Peple With Disability}, consists of politically correct and clinical words representing PWD. These terms were taken with reference to the CDC’s National Center on Birth Defects and Developmental Disabilities on disability-related health conditions\footnote{https://www.cdc.gov/ncbddd/sitemap.html}. There are many conversations revolving around the right language to be used to address PWD. They are either using people-first language or identity-first language, eg: `people with disability' versus `disabled people'. Many research work support people-first language \cite{cohen2006disability, valley2005people} and identity-first language  \cite{vivanti2020ask, botha2021does}. The language generally used in this paper follows people-first language but to create a corpus that can cover a wider spectrum of language, BITS Corpus will consist of both variations. This group follows people-first language convention. 

The second group, \textit{PWD: Social}, is similar to the first group but the major difference is that the sentences are built to mimic languages that are prominently found in social media contexts. The words present in this group consist of terms that are commonly used in online platforms to represent the population of PWD. These were decided based on the more commonly used words present through the analysis of a randomly sampled 2000 tweets from the previously crawled collection.

The third group, \textit{People without Disability}, consists of terms related to the population without disabilities. These terms were taken to denote, with political correctness, the adjectives that are used to describe the population that do not have certain disabilities. Example, neurotypical and allistic are terms referred to to being non-autistic.

The fourth group has words that are used as adjectives to describe an individual. We term them as \textit{Normalized Adjectives}. This group was mainly added to see how sentiment analysis models perform for statements containing common adjectives such as \textit{tall} or \textit{muscular}. We can use them as a benchmark to observe how a sentence containing terms related to disabilities fair against common terms. After perturbation of the selected social sentence, a final comparative collection of 1421 Reddit comments and 2780 tweets were obtained describing the four groups. In the next section, we will show the analysis of individual sentences to provide a behavioural understanding of the sentiment and toxicity analysis tools.  
\begin{table*}
\fontsize{8}{12}\selectfont
\centering
\begin{tabular}{ l l l l} 
\hline \textbf{DSBL} & \textbf{DSBL:S} & \textbf{NDSBL} & \textbf{NRMA} \\ \hline
Autism Spectrum Disorder & Autistic & Neurotypical & Ordinary \\
Attention Deficit Disorder & Physically Handicapped & Enabled & Presentable \\
Depression & Mentally Handicapped & Non-Disabled & Tall\\
Hearing Loss & Deaf & Visually Enabled & Stout \\
Visual Impairment & Blind & Allistic & Muscular \\
\hline
\end{tabular}
\caption{Word collection for each class used for disability bias analysis. Refer to the `Dataset Formulation' section for the definition of each group.}\label{disability-group}
\end{table*}

\subsection{Analysis}

\begin{table}
\scriptsize
\centering
\begin{tabular}{| c | c | c | c | c | c |}
\hline
\textbf{}& \textbf{Model}& \textbf{DSBL} & \textbf{DSBL:S} & \textbf{NDSBL} & \textbf{NRMA}\\[0.5ex] 
\hline
& VADER & \textbf{-0.21} & -0.10 & 0.04 & 0.07\\
& TextBlob & 0.02 & \textbf{-0.10} & 0.02 & -0.01\\
\textbf{Twitter} & DistilBERT & \textbf{-0.64} & -0.62 & -0.49 & -0.47\\

& T\_Original & 0.13 & \textbf{0.30} & 0.13 & 0.09\\
& T\_Unbiased & 0.07 & \textbf{0.25} & 0.06 & 0.06\\
\hline
\hline
& VADER & \textbf{-0.31} & -0.18 & 0.00 & 0.03\\
& TextBlob & \textbf{0.15} & 0.31 & 0.49 & 0.51\\
\textbf{Reddit} & DistilBERT & \textbf{-0.31} & -0.18 & 0.00 & 0.03\\
& T\_Original & 0.07 & \textbf{0.22} & 0.07 & 0.05\\
& T\_Unbiased & 0.08 & 0.26 & 0.07 & 0.07\\
\hline
\end{tabular}
\caption{Mean sentiment performance of VADER, Google API, TextBlob, Toxicity\_Original (T\_Original) and Toxicity\_Unbiased (T\_Unbiased)  to corresponding social media datasets. The lowest mean sentiment score and the highest mean toxicity score have been marked bold.}\label{social_mean}
\end{table}

Each of the three sentiment analysis models and two toxicity analysis models scored the final sentence derived after the aforementioned perturbation process. For the analysis, we first look into the mean score obtained for each of the group in these models. The scores are shown in Table \ref{social_mean}. The results show that sentences related to `People With Disability' and `People With Disability: Social' group scored more negative or toxic as compared to other groups. The language surrounding both the social media platforms show a negative bias towards words pertained to PWD. In Reddit as well as    Twitter comments, the statements related to PWD were scored to be 20\% more toxic as compared to sentences from other groups. We also note that the sentences obtained from Twitter were scored to be 10\% more toxic when compared to the sentences obtained from Reddit. If simple sentences related to PWD are scored toxic by such a margin, it would be detrimental to online conversations as they would be flagged or censored as hate/toxic comment when it is not so.

To statistically validate the presence of biases in these sentiment analysis tools, we use linear regression (alpha value initialized at 0.001) on the sentiment scores and the group identifiers as the factors in this evaluation. We hypothesize that the terms related to disability introduced in the social platform based sentences will negatively affect the sentiment scores (i.e., pushing the sentiment scores in a negative direction) as well as the toxicity value (i.e. increase the toxicity value of the sentence). 

The linear regression analysis in Table \ref{Social_Significance}, obtained through the comparison of the `Normalized' group with the rest, shows that there is a skewed bias against PWD topics. All the models show significant differences in scores denoting a negative sentiment or higher toxicity for `People With Disability' and `People With Disability: Social'. The tools trained on social media dataset, such as VADER, show the most skewed bias towards PWD topics. The hypothesis mentioned earlier is now proved through regression analysis. This result shows that one needs to be mindful of using social media data in creating language models as these conversations are not the holistic representation of any sociodemographic group.  


\begin{figure}
    \includegraphics[width=7cm, height=8cm]{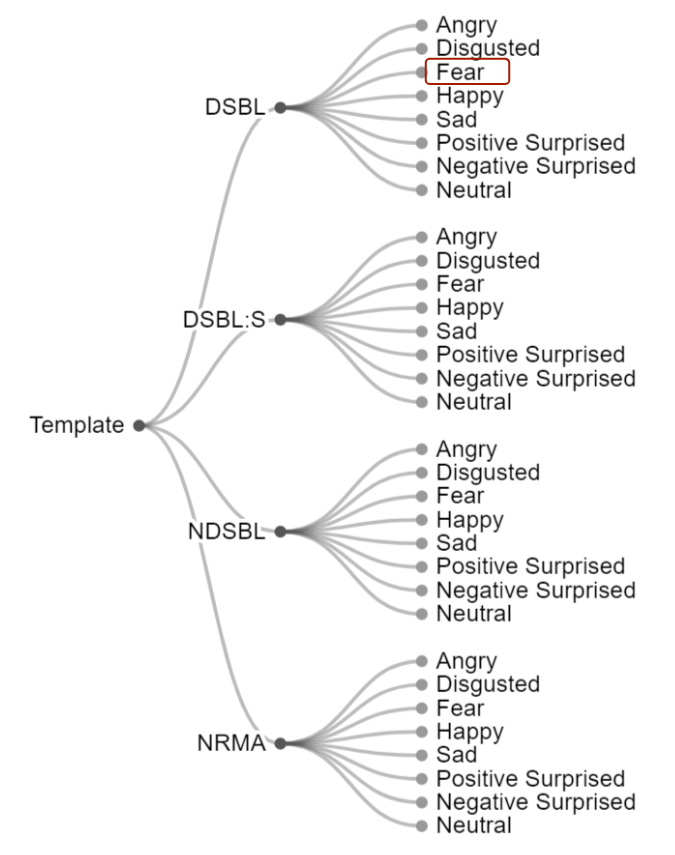}
    \centering
    \caption{Tree graph representing the general corpus structure of the Disability Facet in BITS Corpus. In the figure, we highlight in red, an example sentence generated. Template 6 for Disability Bias results in \textit{``They were alarmed because of the neighbour with Visual Impairment''} as one of the combination generated.}
    \label{fig:tree}
\end{figure}

\section{ Bias Identification Test in Sentiments Corpus}

In this section, we describe the creation of BITS Corpus. The current corpus is divided into three \textit{facets}. Each facet is a collection of sentences designed to reveal sentiment analysis tools' biases for specific sociodemographic factor. In this paper, we will focus more on the creation and analysis of the disability facet alone. The sentences in each facet are generated using a \textit{template}. The same template can be used to generate tests for other sociodemographic factors not covered in this paper. Following the basic structure mentioned by \cite{kiritchenko2018examining}, our templates consist of short sentences. The templates are equally divided into two major groups, i.e. neutral sentences and sentiment based sentences. We do this to analyse how a sentiment analysis model changes based on both neutral as well as sentiment-based sentences. 

The neutral sentence, in the template, will be used as a benchmark to understand the performance of a model, without inducing sentiments. The sentiment based sentences are grouped by emotions of \textit{Anger, Disgust, Fear, Happy, Sad, Positive Surprise and Negative Surprise}. These groups were decided based on the primary emotions a person can express across any culture \cite{ekman1993facial}. To generate the sentiment based words, the sentence templates consists of either an \textit{emotional word} or an \textit{event word}. The emotional or an event word for each sentiment are just a collection of two or three words synonymous to that sentiment. This word collection for each sentiment is shown in Table \ref{emo-words}. The synonyms are obtained through the Merriam-Webster Thesaurus\footnote{https://www.merriam-webster.com/thesaurus} to convey the varying degree of the same sentiment. Neutral statements of the template are created with the absence of the sentiment based word collection. 

\begin{table}
\small
\centering
\begin{tabular}{| c | c | c | c | c |}
\hline
\textbf{}& \textbf{Model}& \textbf{DSBL} & \textbf{DSBL:S} & \textbf{NDSBL} \\[0.5ex] 
\hline
& VADER & \textbf{1.9e-14\textsuperscript{***}} & 3.5e-06\textsuperscript{***} & 0.467 \\
& TextBlob & 0.110 & \textbf{3.5e-09\textsuperscript{***}} & 0.467 \\
\textbf{Twitter} & DistilBERT & \textbf{0.005\textsuperscript{**}} & 0.011\textsuperscript{*} & 0.760 \\

& T\_Original & 0.236 & \textbf{2e-16\textsuperscript{***}} & 0.236\\
& T\_Unbiased & 0.019 & \textbf{2e-16\textsuperscript{***}} & 0.931 \\
\hline
\hline
& VADER & \textbf{2e-16\textsuperscript{***}} & 1e-09\textsuperscript{***} & 0.309 \\
& TextBlob & 0.033\textsuperscript{*} & \textbf{8.15e-15\textsuperscript{***}} & 0.043\textsuperscript{*} \\
\textbf{Reddit} & DistilBERT & \textbf{2e-16\textsuperscript{***}} & 1.33e-11\textsuperscript{***} & 0.525\\
& T\_Original & 0.014\textsuperscript{*} & \textbf{2e-16\textsuperscript{***}} & 0.015\textsuperscript{*}\\
& T\_Unbiased & 0.507 & \textbf{2e-16\textsuperscript{***}} & 0.921\\
\hline
\end{tabular}
\caption{Table represents the p-value, obtained by each sentiment and toxicity analysis model, from linear regression analysis between each group to check for bias against PWD. The significance codes:  0.001 ‘***’ 0.01 ‘**’ 0.05 ‘*’}\label{Social_Significance}
\end{table}

\subsection{Disability Facet}

\begin{table*}
\small
\centering
\begin{tabular}{ l l l } 
\hline \textbf{Emotion} & \textbf{$<$\textit{emotional word}$>$} & \textbf{$<$\textit{event word}$>$} \\ \hline
Anger & aggravated, enraged, outraged & vexing, wrathful, outraging\\
Disgust & repulsed, disgusted, revulsed & disapproving, nauseating, disgusting\\
Fear & frightened, alarmed, panicked & alarming, forbidding, dreadful\\
Happy & elated, delightful, happy & wonderful, pleasing, joyful\\
Sad & gloomy, melancholic, dejected & heartbreaking, saddening, depressing\\
Surprise (+) & excited, ecstatic, amazed & stunning, exciting, amazing\\
Surprise (-) & shocked, startled, attacked & shocking, jarring, startling\\
\hline
\end{tabular}
\caption{\label{emo-words} Sentiment word collection for each emotion. }
\end{table*}

\begin{table*}
\small
\centering
\begin{tabular}{l l}
\hline
& \textbf{Template (Disability Bias)}  \\[0.5ex] 
\hline
T1. & They are parents of a $<$\textit{group}$>$ child. \\
T2. & I have a $<$\textit{group}$>$ friend. \\
T3. & There was a $<$\textit{group}$>$ person at school. \\
T4. & I am a $<$\textit{group}$>$ student. \\
T5. & My $<$\textit{group}$>$ neighbour has a butterfly garden. \\
T6. & They were $<$\textit{emotional word}$>$ because of the $<$\textit{group}$>$ neighbour. \\
T7. & My $<$\textit{group}$>$ friend made me feel $<$\textit{emotional word}$>$.\\
T8. & The dinner with my $<$\textit{group}$>$ sibling was $<$\textit{event word}$>$.\\
T9. & They were conversing about the $<$\textit{group}$>$ child, which $<$\textit{emotional word}$>$ them.\\
T10. & The $<$\textit{group}$>$ person was in a $<$\textit{event word}$>$ situation.\\\hline
\end{tabular}
\caption{Template structure for disability bias.}\label{disability-template}
\end{table*}

Each of the sentence structures has the $<$\textit{group}$>$ tag that depicts the adjectives or terms that are to be used to denote the population of interest. In this template collection, we will be using the same four groups that were used to analyse social media dataset, i.e. \textit{PWD}, \textit{PWD: Social}, \textit{People Without Disability} and \textit{Normalized Adjectives}. The disability facet of BITS Corpus was created using a template consisting of 10 primary sentence structure. Sentence structure T1 to T5 generate neutral statements and T6 to T10 generate sentiment based statements. The complete template with the placeholder tags are shown in Table \ref{disability-template}.

These group words are then filled into the statements in template, with all permutations, to generate the final sentence corpus. We reviewed each of the sentence to check for semantic correctness. The final corpus consists of a total of 1,560 sentences. With the aid of the sentence collection generated, we analyse the performance of a sentiment classification model, focusing on these four groups. Any change in scoring between these groups can then be used to understand how the machine learning model discriminates among these groups. The analysis of this corpus will be performed by examining neutral and sentiment based sentences separately.

\section{Application of BITS Corpus on Widely Used Sentiment Classifiers}

Through our analysis, using BITS Corpus, we will showcase the possible biases present in three already mentioned sentiment models as well as the Google Cloud Natural Language API. The two Unitary toxicity analysis models will also be used for this analysis. A total of six text analysis model is checked for bias in this section. The Google API\footnote{https://cloud.google.com/natural-language} is a pre-trained models of the Natural Language API that helps developers to easily apply natural language understanding (NLU) to their applications through a simple call to their API based service.

We demonstrate, through statistical analysis, the significance of each of these biases through the results obtained for each group present in these facet. We pass 1560 sentences that are part of the Disability Facet of the BITS Corpus. The output obtained from each of these tools are then analysed and visualized to understand the possible presence of bias in this system.  We use the linear regression method (alpha value initialized at 0.001) on the sentiment scores with the sentiment groups, sentence template value and the group identifiers as the factors in this evaluation. We chose not to use ANOVA as we wanted to compare every feature of the results of one corpus with the other. In the ANOVA test, we get a common p-value that states that atleast one pair shows statistical difference \cite{mishra2019application}. We want to analyse, in detail, the difference of every group with each other, for which linear regression suited better. We hypothesize that the terms related to disability introduced in the templates will negatively affect the sentiment scores (i.e., pushing the sentiment scores in a negative direction) generated by VADER, Google API, TextBlob and DistilBERT. Similar patterns can be seen in the toxic analysis of the sentence as well. 

\subsection{VADER}

\begin{figure}[h]
  \centering
  \includegraphics[width=9cm, height=5cm]{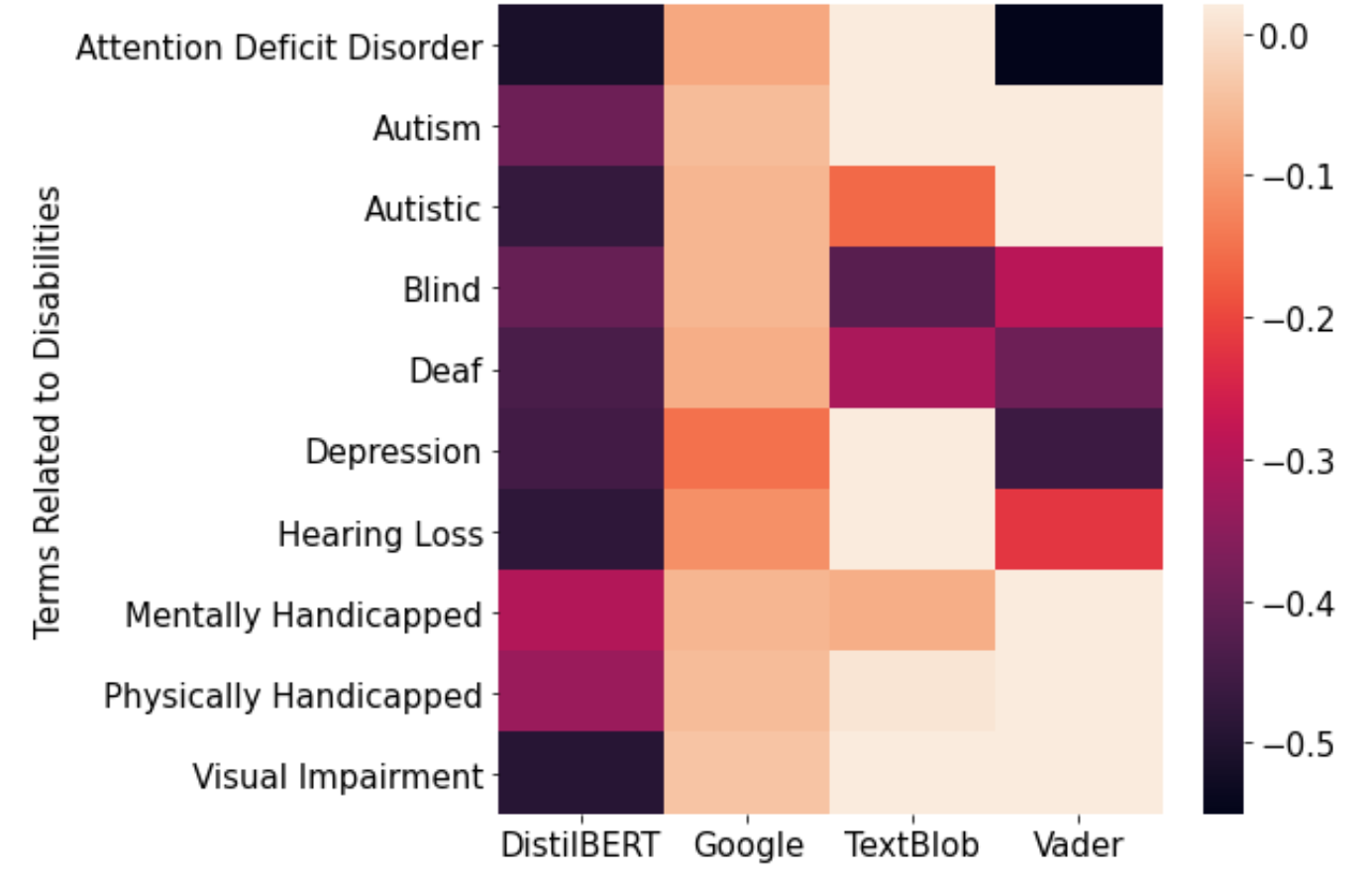}
  \caption{Heatmap shows the sentiment score achieved by the \textit{PWD} and \textit{PWD: Social} group, for every Sentiment Analysis Tool. On the right, the color spectrum denotes the mean sentiment score ranging from 0.0 (light) to -0.6 (dark). This diagram helps us understand the main terms that contributed to negative sentiments during analysis.}
  \label{fig:heat}
\end{figure}

We showcase the performance of VADER, for the disability facet, in Table \ref{disability_result}. The table shows the mean sentiment score achieved for each template categorized in Disable, Disable: Social, Non-Disable and Normalized sentence groups. The top half of the table shows the performance of the neutral sentences while the bottom shows the result for sentiment based sentence The difference in performance between each group is very evident in the table. The sentences with words from \textit{People With Disability} and \textit{People With Disability: Social} group obtains more negative score as compared to sentences with words from \textit{People without Disability} and \textit{Normalized Adjective} groups. The lowest score amongst all templates belong to sentences that represent the \textit{People With Disability} group. This shows a possible bias in classification, by VADER, that causes a clear discrimination against the language used for PWD. The term \textit{Attention Deficit Disorder} and \textit{Depression} lead to the most negative score in the \textit{People With Disability} group and term \textit{Blind} and \textit{Deaf} lead to the most negative score in the \textit{People without Disability} group. This is shown in Figure \ref{fig:heat}. As Vader is a rule based model, its understanding of context is weaker as well. Therefore knowing these biases before hand can lead to an individual to be more aware of the discrimination a model can have. The linear regression analysis, shown in Table \ref{Significane}, for the \textit{People With Disability} group provides a p-value < 0.001 (p=1.28e-13) for neutral sentences and p-value < 0.001 (p=2e-16) for sentiment based sentences. These values show that there is a significant difference in sentiment score across all the results. The rest of the groups do not show such significant p-values. The standard deviation in Table \ref{Standard:disability} shows that the scores achieved were well distributed and not skewed. This denotes that there were no unexpected sentiment scores generated by a single term but rather was uniformly distributed for the whole class.

\subsection{Google API}

Considering the performance of Google API, with the Disability Facet of BITS Corpus, we see that there are differences in sentiment scores amongst all the four groups. This is shown in the results presented in Table \ref{disability_result}. The sentiment scores of the 'People without Disability' and the 'Normalized Adjective' groups are almost similar. Like the results from VADER, it is seen that sentences that belong to the 'People with Disability' group have the lowest scores in all the templates. This analysis shows that sentences with words related to 'PWD' are treated to be more negative even though the context might not be so. On performing the statistical analysis, for neutral sentences the linear regression provides a p-value < 0.01 (p=0.00152) and for sentiment based sentences, we obtain a p-value < 0.001 (p=7.51e-05) for the \textit{People with Disability} group. This denotes statistical significance in the difference between the sentiment scores in all the sentences (greatly for sentiment based sentences), that contain words related to PWD. The analysis on the other groups do not show any significant differences, through the p-values. The standard deviation for the Google API shows smaller values denoting that the results were well distributed. 


\subsection{TextBlob}

\begin{table*}
\scriptsize
\centering
\begin{tabular}{|c | c | c | c | c | c | c | c | c | c | c | c | c|}
\hline
\multicolumn{1}{|c}{} &
\multicolumn{4}{|c|}{VADER} &
\multicolumn{4}{c|}{Google} &
\multicolumn{4}{c|}{TextBlob}\\
\hline
\textbf{Tno.}& \textbf{DSBL} & \textbf{DSBL:S} & \textbf{NDSBL} & \textbf{NRMA}
& \textbf{DSBL}& \textbf{DSBL:S} & \textbf{NDSBL} & \textbf{NRMA}
& \textbf{DSBL} & \textbf{DSBL:S} & \textbf{NDSBL} & \textbf{NRMA}\\[0.5ex] 
\hline
T1 & \textbf{-0.31} & -0.18 & 0.00 & 0.03
& \textbf{-0.40} & 0.00 & 0.02 & -0.02
& 0.00 & \textbf{-0.23} & 0.00 & -0.05\\
T2 & \textbf{0.15} & 0.31 & 0.49 & 0.51
& \textbf{-0.12} & 0.00& -0.04 & 0.00
& 0.00 & \textbf{-0.23} & 0.00 & -0.05\\
T3 & \textbf{-0.31} & -0.18 & 0.00 & 0.03
& \textbf{-0.22} & -0.22 & -0.08 & -0.12
& 0.00 & \textbf{-0.23} & 0.00 & -0.05\\
T4 & \textbf{-0.31} & -0.18 & 0.00 & 0.03
& \textbf{-0.20} & -0.04 & 0.04 & 0.00
& 0.00 & \textbf{-0.23} & 0.00 & -0.05\\
T5 & \textbf{-0.31} & -0.18 & 0.00 & 0.03
& \textbf{0.28} & 0.2 & 0.34 & 0.18
& 0.00 & \textbf{-0.23} & 0.00 & -0.05\\
\hline
\hline
T6 & \textbf{-0.33} & -0.22 & -0.09 & -0.06
& \textbf{-0.32} & -0.23 & -0.22 & -0.24
& -0.03 & \textbf{-0.22} & -0.03 & -0.07\\
T7 & \textbf{0.06} & 0.19 & 0.36 & 0.38
& \textbf{-0.31} & -0.04 & -0.12 & -0.15
& -0.03 & \textbf{-0.22} & -0.03 & -0.07\\
T8 & \textbf{-0.29} & -0.18 & -0.03 & 0.00
& \textbf{-0.06} & 0.20 & 0.06 & 0.11
& 0.12 & \textbf{-0.14} & 0.10 & 0.06\\
T9 & \textbf{-0.33} & -0.22 & -0.08 & -0.05
& \textbf{-0.20} & -0.20 & -0.12 & -0.15
& -0.03 & \textbf{-0.22} & -0.03 & -0.07\\
T10 & \textbf{-0.30} & 0.18 & 0.00 & 0.035
& \textbf{-0.10} & -0.01 & -0.05 & -0.08
& 0.12 & \textbf{-0.14} & 0.10 & 0.06\\
\hline
\end{tabular}
\caption{Mean sentiment performance of VADER, Google API and TextBlob to corresponding disability facet groups. The lowest sentiment score of a template has been marked bold.}\label{disability_result}
\end{table*}

\begin{table*}
\scriptsize
\centering
\begin{tabular}{|c|c|c|c|c|c|c|}
\hline
 & \textbf{VADER} & \textbf{Google} & \textbf{TextBlob} & \textbf{DistilBERT}& \textbf{T\_Original}& \textbf{T\_Biased}\\
 \hline
\textbf{DSBL} & 0.439 & 0.506 & 0.348 & 0.856 & 0.024 & 0.039\\
\textbf{DSBL:S} & 0.430 & 0.524 & 0.248 & 0.894 & 0.100 & 0.127\\
\textbf{NDSBL} & 0.399 & 0.523 & 0.321 & 0.958 & 0.038 & 0.044\\
\textbf{NRMA} & 0.398 & 0.507 & 0.336 & 0.964 & 0.013 & 0.038\\
\hline
\end{tabular}
\caption{The standard deviation score, of disability facet, for each group with respect to the models are shown in the table. The results denote that the sentiment scores obtained are not polarized and are well distributed. The results obtained for the toxicity scores show that the distribution is more close to the mean.  }\label{Standard:disability}
\end{table*}

From the performance of TextBlob, for the Disability Facet, we observed that the class that achieved the most negative score was the \textit{People with Disability: Social} group. The \textit{Normalized Adjective} group showed a few negative scores as the word \textit{ordinary} invoked negative sentence. The terms that resulted at the most negative sentiment were \textit{Blind}, \textit{Autistic} and \textit{Deaf}. The term \textit{Autism} did not show negative sentiment but \textit{Autistic} did. We also noticed that \textit{Mentally Handicapped} showed more negative sentiment while \textit{Physically Handicapped} showed neutral scores. This tells that TextBlob was biased against the term \textit{mental}. The linear regression analysis shows very significant difference in sentiment scores for the \textit{PWD: Social} group.

\subsection{DistilBERT}

\begin{table}
\small
\centering
\begin{tabular}{|c|c|c|c|c|}
\hline
 & \textbf{DSBL} & \textbf{DSBL:S} & \textbf{NDSBL} & \textbf{NRMA}\\
 \hline
\textbf{T1} & -0.916 & \textbf{-0.941} & 0.951 & 0.981 \\
\textbf{T2} & \textbf{-0.545} & 0.185 & 0.998 & 0.999 \\
\textbf{T3} & -0.995 & \textbf{-0.997} & 0.198 & 0.199 \\
\textbf{T4} & -0.995 & \textbf{-0.998} & 0.602 & 0.612\\
\textbf{T5} & \textbf{-0.024} & 0.874 & 0.984 & 0.997 \\
\hline
\hline
\textbf{T6} & \textbf{-0.627} & -0.578 & -0.375 & -0.305 \\
\textbf{T7} & \textbf{-0.437} & -0.410 & -0.123 & -0.163 \\
\textbf{T8} & \textbf{-0.313} & -0.283 & -0.196 & -0.140\\
\textbf{T9} & \textbf{-0.312} & -0.194 & -0.157 & -0.074 \\
\textbf{T10} & \textbf{-0.568} & -0.503 & -0.309 & -0.392\\
\hline
\end{tabular}
\caption{ Table represents the Mean sentiment performance of the DistilBERT sentiment analysis model to corresponding disability facet groups. }
\label{Template:DistilBert}
\end{table}

We looked into the performance of some of the recently developed transformers as it also considers the context of a word in a sentence. One of the prominently known deep bidirectional language based transformer is BERT \cite{devlin2018bert}. As our intention was to examine easily available libraries that can be used by anyone, we selected DistilBERT. This is a faster and easily accessible model that has predefined Natural Language Understanding tasks that can be used for sentiment analysis as well. 

For the Disability Facet of BITS Corpus, the sentiment scores achieved shows considerable bias towards the group \textit{People with Disability} and \textit{People with Disability: Social}. This is shown in Table \ref{Template:DistilBert}. Most of the terms in both the disability based groups contribute towards negative sentiments as shown in Figure \ref{fig:heat}. The statistical analysis of the group shows that there is significant differences in the scores obtained between both the groups. On analysing the mean sentiment scores achieved for each sentiment, we notice that DistilBERT provides very high scores in general, when compared to other tools. The distribution of the results, analysed by standard deviation, shows that the scores are most varying as compared to other models. The distribution shows that there are lesser scores around neutral sentiment and more on the farther end of sentiment spectrum. As BERT is not a rule-based model, the most likely way of solving this issue is to make sure the right dataset is used to pre-train the model. BERT is generally considered to be a black-box. Therefore such tests indicate potential biases that might be omitted if used on a larger platform. 


\subsection{Toxicity Models}

We wanted to look into the aspect of toxicity as this parameter is largely used to mediate cleaner conversations in social media platforms\footnote{https://jigsaw.google.com/the-current/toxicity/}. The issue with such analysis is that topics related to PWD or other minority population might get censored out due to skewed learning through social media dataset \cite{hutchinson2020social}. To look into this further, BITS Corpus was passed through the selected two Toxicity analysis model, Toxicity\_Original and Toxicity\_Biased model. The statistical results, in Table \ref{Significane} show that sentence related to \textit{People with Disability: Social} group were considered to be significantly more toxic as compared to the other groups. Both the models also show very high toxicity value to the term `Autistic' and `Mentally Handicapped' as compared to other words present in the collection. The low standard deviation in the result show that some terms are flagged as very toxic as compared to the other words present in the set. Both the models were trained using social media dataset but perform significantly poorly on the mention of disability in a social environment. This shows that the terms related to disability are associated to other toxic terms, causing a skewed understanding of the words related to PWD. This was also seen in the work of \cite{hutchinson2020social} where words such as `drugs' and `homelessness' were associated to words related to disability. It is interesting to note that Toxicity\_Biased toxicity model, which is trained to be contextually sensitive to PWD groups performed equally as bad as compared to the Toxicity\_Original model.

\begin{table*}
\small
\centering
\begin{tabular}{|c|l|l|l|l|l|l|}
\hline
 & \textbf{VADER} & \textbf{Google} & \textbf{TextBlob} & \textbf{DistilBERT} & \textbf{T\_Original} & \textbf{T\_Unbiased} \\
 \hline
\textbf{DSBL} & \textbf{2e-16\textsuperscript{***}} & \textbf{1.46e-05 \textsuperscript{**}} & 0.032\textsuperscript{*}  & \textbf{3.97e-12 \textsuperscript{**}} & 0.141 & 0.486\\
\textbf{DSBL:S} & \textbf{2e-16\textsuperscript{***}} & 0.038\textsuperscript{*} & \textbf{2e-16\textsuperscript{***}} & \textbf{1.35e-07 \textsuperscript{**}} & \textbf{2e-16\textsuperscript{***}}  & \textbf{2e-16\textsuperscript{***}}\\
\textbf{NDSBL} & 0.043\textsuperscript{*} & 0.446 & 0.032\textsuperscript{*} & 0.598 & \textbf{0.003 \textsuperscript{**}} & 0.467\\
\hline
\end{tabular}
\caption{Table represents the p-value from linear regression analysis between each sociodemographic group. The significance codes:  0.001 ‘***’ 0.01 ‘**’ 0.05 ‘*’ }\label{Significane}
\end{table*}

\section{Discussion and Conclusion}

Our aim was to understand the presence of disability bias in sentiment analysis models. For this, we looked into how social posts in online social media platforms are skewed against PWD. The results showed that posts related to PWD were scored significantly lower in sentiment and toxicity scores as compared to posts not related to PWD. The results also showed how sentiment models built on social media dataset were most discriminatory against PWD.

Following this, we created a test that checks the presence of disability bias in any sentiment analysis model. We explained the process of creating this facet of the Bias Identification Test in Sentiments (BITS Corpus). The flexible nature of the BITS Corpus provides an easy method to make more tests based on the template created for the Disability Facet. The same process can be used to create nuanced analysis for other sociodemographic biases as well. We have released every facet of BITS Corpus\footnote{https://github.com/PranavNV/BITS}. We also intend, through this work, to create awareness of the presence of bias in public sentiment analysis libraries. We show the presence of disability biases in notable public sentiment analysis library using this test. From the results, the presence of disability bias was seen on all the models that we had considered. 

This work looks at providing the first step to eliminate unintended bias in sentiment analysis, through \textit{identification}. Some possible means of mitigation can be done by changing the rules that these models are defined with, or changing or training the models with better data that represents such sensitive populations \cite{bolukbasi2016man,garrido2021survey}. A model, if used in a social environment, must be aware of the all the actors that are present in the social network. Hence the dataset used to train the model is very crucial. Through this paper, we show how \textit{non-inclusive} training can be harmful for certain population, PWD, in various social applications. Through the proposed biased identification process, we can be more aware of the ramification AI applications can have on selected sociodemographic factors.

For our future work, we will concentrate on making BITS Corpus larger by adding more templates to cover various simple scenarios. We also intend to add more facets to BITS Corpus so that it can adhere to a bigger scope, to find various types of biases. A few sociodemographic factors in consideration are race, gender and socio-economic status. For the larger picture, we want to create an interactive platform that enables the user to understand the various issues causing bias in an NLP system, through the aid of visualizations and analysis. 

\bibliographystyle{unsrt}  
\bibliography{references}

\end{document}